\documentclass[letterpaper, 10 pt, conference]{ieeeconf}  

\IEEEoverridecommandlockouts                              

\usepackage[dvipsnames]{xcolor}
\usepackage{graphics} 
\usepackage{epsfig} 
\usepackage{times} 
\usepackage{amsmath} 
\usepackage{amssymb}  
\usepackage[caption=false]{subfig}
\usepackage{multirow}
\usepackage{placeins}

\makeatletter
\let\NAT@parse\undefined
\makeatother
\usepackage[colorlinks,allcolors=black,urlcolor=blue]{hyperref}

\graphicspath{{./imgs/}{./imgs/evaluations/abs_pose/}{./imgs/evaluations/real_recon_tank/}}

\renewcommand{\d}[1]{\mbox{\boldmath$#1$}}

\newcommand{\trans}[0]{^{\sf T}}

\newcommand{\q}[1]{{{\bf #1}}}

\newcommand{\mq}[1]{{\q #1}}

\newcommand{\mC}{\mathchoice {\setbox0=\hbox{$\displaystyle\rm
C$}\hbox{\hbox to0pt{\kern0.4\wd0\vrule height0.9\ht0\hss}\box0}}
{\setbox0=\hbox{$\textstyle\rm C$}\hbox{\hbox
to0pt{\kern0.4\wd0\vrule height0.9\ht0\hss}\box0}}
{\setbox0=\hbox{$\scriptstyle\rm C$}\hbox{\hbox
to0pt{\kern0.4\wd0\vrule height0.9\ht0\hss}\box0}}
{\setbox0=\hbox{$\scriptscriptstyle\rm C$}\hbox{\hbox
to0pt{\kern0.4\wd0\vrule height0.9\ht0\hss}\box0}}}

\newcommand{\mG}{\mathchoice {\setbox0=\hbox{$\displaystyle\rm
G$}\hbox{\hbox to0pt{\kern0.4\wd0\vrule height0.9\ht0\hss}\box0}}
{\setbox0=\hbox{$\textstyle\rm G$}\hbox{\hbox
to0pt{\kern0.4\wd0\vrule height0.9\ht0\hss}\box0}}
{\setbox0=\hbox{$\scriptstyle\rm G$}\hbox{\hbox
to0pt{\kern0.4\wd0\vrule height0.9\ht0\hss}\box0}}
{\setbox0=\hbox{$\scriptscriptstyle\rm G$}\hbox{\hbox
to0pt{\kern0.4\wd0\vrule height0.9\ht0\hss}\box0}}}

\newcommand{\mJ}{\mathchoice {\setbox0=\hbox{$\displaystyle\rm
J$}\hbox{\hbox to0pt{\kern0.4\wd0\vrule height0.9\ht0\hss}\box0}}
{\setbox0=\hbox{$\textstyle\rm J$}\hbox{\hbox
to0pt{\kern0.4\wd0\vrule height0.9\ht0\hss}\box0}}
{\setbox0=\hbox{$\scriptstyle\rm J$}\hbox{\hbox
to0pt{\kern0.4\wd0\vrule height0.9\ht0\hss}\box0}}
{\setbox0=\hbox{$\scriptscriptstyle\rm J$}\hbox{\hbox
to0pt{\kern0.4\wd0\vrule height0.9\ht0\hss}\box0}}}

\newcommand{\mO}{\mathchoice {\setbox0=\hbox{$\displaystyle\rm
O$}\hbox{\hbox to0pt{\kern0.4\wd0\vrule height0.9\ht0\hss}\box0}}
{\setbox0=\hbox{$\textstyle\rm O$}\hbox{\hbox
to0pt{\kern0.4\wd0\vrule height0.9\ht0\hss}\box0}}
{\setbox0=\hbox{$\scriptstyle\rm O$}\hbox{\hbox
to0pt{\kern0.4\wd0\vrule height0.9\ht0\hss}\box0}}
{\setbox0=\hbox{$\scriptscriptstyle\rm O$}\hbox{\hbox
to0pt{\kern0.4\wd0\vrule height0.9\ht0\hss}\box0}}}

\newcommand{\mQ}{\mathchoice {\setbox0=\hbox{$\displaystyle\rm
Q$}\hbox{\raise 0.15\ht0\hbox to0pt{\kern0.4\wd0\vrule
height0.8\ht0\hss}\box0}}{\setbox0=\hbox{$\textstyle\rm Q$}\hbox{\raise
0.15\ht0\hbox to0pt{\kern0.4\wd0\vrule height0.8\ht0\hss}\box0}}
{\setbox0=\hbox{$\scriptstyle\rm Q$}\hbox{\raise 0.15\ht0\hbox
to0pt{\kern0.4\wd0\vrule height0.7\ht0\hss}\box0}}{\setbox0=
\hbox{$\scriptscriptstyle\rm Q$}\hbox{\raise 0.15\ht0\hbox
to0pt{\kern0.4\wd0\vrule height0.7\ht0\hss}\box0}}}

\newcommand{\mS}{\mathchoice
{\setbox0=\hbox{$\displaystyle     \rm S$}\hbox{\raise0.5\ht0\hbox
to0pt{\kern0.35\wd0\vrule height0.45\ht0\hss}\hbox
to0pt{\kern0.55\wd0\vrule height0.5\ht0\hss}\box0}}
{\setbox0=\hbox{$\textstyle        \rm S$}\hbox{\raise0.5\ht0\hbox
to0pt{\kern0.35\wd0\vrule height0.45\ht0\hss}\hbox
to0pt{\kern0.55\wd0\vrule height0.5\ht0\hss}\box0}}
{\setbox0=\hbox{$\scriptstyle      \rm S$}\hbox{\raise0.5\ht0\hbox
to0pt{\kern0.35\wd0\vrule height0.45\ht0\hss}\raise0.05\ht0\hbox
to0pt{\kern0.5\wd0\vrule height0.45\ht0\hss}\box0}}
{\setbox0=\hbox{$\scriptscriptstyle\rm S$}\hbox{\raise0.5\ht0\hbox
to0pt{\kern0.4\wd0\vrule height0.45\ht0\hss}\raise0.05\ht0\hbox
to0pt{\kern0.55\wd0\vrule height0.45\ht0\hss}\box0}}}
\newcommand{\mT}{\mathchoice {\setbox0=\hbox{$\displaystyle\rm
T$}\hbox{\hbox to0pt{\kern0.3\wd0\vrule height0.9\ht0\hss}\box0}}
{\setbox0=\hbox{$\textstyle\rm T$}\hbox{\hbox
to0pt{\kern0.3\wd0\vrule height0.9\ht0\hss}\box0}}
{\setbox0=\hbox{$\scriptstyle\rm T$}\hbox{\hbox
to0pt{\kern0.3\wd0\vrule height0.9\ht0\hss}\box0}}
{\setbox0=\hbox{$\scriptscriptstyle\rm T$}\hbox{\hbox
to0pt{\kern0.3\wd0\vrule height0.9\ht0\hss}\box0}}}
\newcommand{\mU}{\mathchoice {\setbox0=\hbox{$\displaystyle\rm
U$}\hbox{\hbox to0pt{\kern0.4\wd0\vrule height0.9\ht0\hss}\box0}}
{\setbox0=\hbox{$\textstyle\rm U$}\hbox{\hbox
to0pt{\kern0.4\wd0\vrule height0.9\ht0\hss}\box0}}
{\setbox0=\hbox{$\scriptstyle\rm U$}\hbox{\hbox
to0pt{\kern0.4\wd0\vrule height0.9\ht0\hss}\box0}}
{\setbox0=\hbox{$\scriptscriptstyle\rm U$}\hbox{\hbox
to0pt{\kern0.4\wd0\vrule height0.9\ht0\hss}\box0}}}

\title{\LARGE \bf
Refractive COLMAP: Refractive Structure-from-Motion Revisited
}

\author{Mengkun She and Felix Seegr\"aber and David Nakath and Kevin K\"oser
\thanks{This work was supported by the German Research Foundation (Deutsche Forschungsgemeinschaft, DFG) Projektnummer 396311425, through the Emmy Noether Program}
\thanks{The authors are with the Department of Computer Science, Christian-Albrechts-University of Kiel, Neufeldtstraße 6, 24118 Kiel, Germany
       {\tt\small \{mshe,fse,dna,kk\}@informatik.uni-kiel.de}}%
}

\begin{document}

\maketitle
\thispagestyle{empty}
\pagestyle{empty}

\begin{abstract} 
In this paper, we present a complete refractive Structure-from-Motion (RSfM) framework for underwater 3D reconstruction using refractive camera setups (for both, flat- and dome-port underwater housings). 
Despite notable achievements in refractive multi-view geometry over the past decade, a robust, complete and publicly available solution for such tasks is not available at present, and often practical applications have to resort to approximating refraction effects by the intrinsic (distortion) parameters of a pinhole camera model.
To fill this gap, we have integrated refraction considerations throughout the entire SfM process within the state-of-the-art, open-source SfM framework COLMAP. 
Numerical simulations and reconstruction results on synthetically generated but photo-realistic images with ground truth validate that enabling refraction does not compromise accuracy or robustness as compared to in-air reconstructions. 
Finally, we demonstrate the capability of our approach for large-scale refractive scenarios using a dataset consisting of nearly 6000 images. 
The implementation is released as open-source at: \url{https://cau-git.rz.uni-kiel.de/inf-ag-koeser/colmap_underwater}.
\end{abstract}

\section{INTRODUCTION}

\begin{figure}
	\centering
	\subfloat[UWPinhole]{
	\includegraphics[width=0.9\columnwidth]{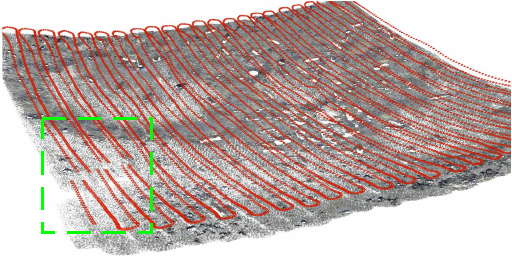}	
	}\\
	\subfloat[RSfM]{
	\includegraphics[width=0.9\columnwidth]{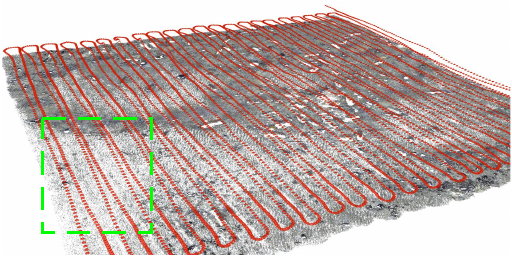}	
}
	\caption{Results of the reconstruction on a rendered large-scale AUV-based seafloor mapping dataset containing 5740 refractive flat-port images. \textbf{Top}: Using the perspective camera model underwater creates a curved seafloor reconstruction. \textbf{Bottom}: Our proposed RSfM.}
        \label{fig:rerender_auv}
\end{figure}

Simultaneous Localization and Mapping (SLAM) as well as Structure-from-Motion (SfM) are key technologies for inferring maps or 3D shapes from images. 
Their application in the underwater domain enables exploration of geological or archaeological sites on the seafloor, mapping or monitoring offshore installations, deposited munitions, or biological habitats, and visually aided autonomous underwater navigation in general.
To protect cameras from water and high pressure in the ocean, they are enclosed in waterproof pressure housings and observe the environment through a transparent window, typically with a planar or spherical shape. 
Light rays from the underwater scene change direction when they travel through these interfaces in a non-orthogonal manner, leading to distortion in the acquired images.
Although refraction is depth-dependant, in the past refraction effects have often been addressed by approximating the entire camera system, including the glass port, as a perspective camera \cite{shortis2015_unwcalib}. 
This enables the use of standard 3D reconstruction software such as COLMAP \cite{schonberger2016structure} and Agisoft Metashape. 
Throughout this work, we refer to this approach as \textbf{UWPinhole}.
However, this approximation is suitable only for certain refractive camera configurations and pre-defined working distances \cite{rofallski2022investigating}, and absorption of distance-dependent refraction into pinhole intrinsics can introduce bias and inconsistencies for large-scale reconstructions (see e.g. Fig. \ref{fig:rerender_auv}).
As an alternative, underwater camera systems can be explicitly modeled with additional physical parameters describing the properties of the housing interface \cite{treibitz_08-flatRefractive,Telem_2010-underwaterCalib,jordt_2016_refractive,she2021refractive}. 
While exact refraction modeling closely resembles physical effects, it invalidates classical pinhole-based multi-view geometry methods.
Integrating these additional physical parameters into SfM therefore remains challenging.

Over the past decade, several solutions have been proposed to address various aspects of the RSfM problem, such as refractive calibration \cite{agrawal2010analytical,JordtSedlazeck_2012RefCalibUnw,she2019adjustment}, refractive motion estimation \cite{chadebecq2019refractive,elnashef2023three,Li_08GeneralEEstHartley,hu2023refractive}, or even partial RSfM system demonstrators \cite{Jordt-Sedlazeck_2013RefSfM,chadebecq2017refractive,elnashef2023drift}, however, limited to flat-ports, lacking open-source implementations, or demonstrated only on a small set of images. 
COLMAP \cite{schonberger2016structure} is widely recognized as a state-of-the-art open-source incremental SfM framework, upon which many downstream tasks like dense Multi-View Stereo \cite{schonberger2016pixelwise} or NeRF \cite{mildenhall2021nerf} depend.
Due to the lack of a suitable underwater alternative, the UWPinhole approximation often remains the commonly applied method in practice \cite{rofallski2022investigating}, and it is even sometimes considered as a reference ground truth in the literature \cite{elnashef2023drift,billings2022hybrid}.
Hence, there remains a need for a complete, open-source, general refractive Structure-from-Motion solution that is proven to be robust, accurate, and capable of handling a large number of images.

In this work, we make the following contributions:
\begin{itemize}
	\item Integration of refraction into COLMAP, supporting generalized refractive camera setups with auto-optimizing the refractive parameters in the reconstruction.
	\item A robust relative pose estimation approach for geometric verification and SfM initialization.
	\item Extensive evaluations on the overall performance of the RSfM pipeline under various refractive camera setups.
\end{itemize} 

\section{Related Work}

\textbf{Refractive Camera Modeling.} Grossberg et al. \cite{Grossberg_05-RaxelImagingModel} and Sch\"ops et al. \cite{schops2020having} utilize a generic ray-based camera model, which directly associates rays from the scene with the image coordinates. In principle, such models could also be used to encode refraction.
In a more specific model for flat glass windows, Treibitz et al. \cite{treibitz_08-flatRefractive} explicitly represent the flat-port interface with a plane and analyze the behavior of rays.
Agrawal et al. \cite{Agrawal_2012-UnwCalib} extend this model to a more general case involving tilted multi-layers interface and demonstrate that the system is an axial camera model. 
Jordt et al. \cite{JordtSedlazeck_2012RefCalibUnw} propose a more comprehensive calibration approach for such systems.
Telem et al.  \cite{Telem_2010-underwaterCalib} propose a varifocal model in which a feature-dependent focal length correction factor is applied to maintain the co-linearity of the ray.
On the other hand, robots for deeper waters are often equipped with spherical glass windows (dome-ports), because they are mechanically much more stable for high water pressures and allow a large field of view.
Additionally, refraction can be avoided if a pinhole camera is perfectly centered within the dome \cite{kunz2008hemispherical,she2019adjustment}. 
However, in practice, de-centering the camera results in behavior akin to an axial camera model \cite{she2021refractive}, similar in spirit to flat-port refraction, but at the sphere.
Nevertheless, 3D reconstruction using non-central camera models requires additional effort.
For special cases, a straightforward approach to avoid addressing this is to undistort refraction before reconstruction. 
This can be achieved by constructing a look-up table using the Pinax model \cite{LUCZYNSKI20179} to map refracted image points back to un-refracted positions. 
However, this technique requires a small camera-to-interface distance in the order of millimeters, assumes a fixed scene distance.
Moreover, the fixed look-up table does not allow refining the refractive calibration during bundle adjustment.

\textbf{Refractive SfM.} Several works exist on SfM using general camera models, such as those by Sturm et al. \cite{sturm2005multi,sturm_06-GenericCameras,Ramalingam_2006genericSfMFramework}.
However, it has been discussed that this model is particularly sensitive to noise. 
Chari et al. \cite{chari2009refractiveplane} provide theoretical insights into multi-view geometry under planar refraction, although without numerical evaluations. 
Jordt-Sedlazeck et al. \cite{Jordt-Sedlazeck_2013RefSfM} is considered as the first approach that tackles the entire SfM problem for underwater imaging. 
Nevertheless, it is only demonstrated on a small-scale scene. 
Elnashef et al. \cite{elnashef2023drift} derive a differential motion model for an axial formulation of the continuous egomotion and propose a visual odometry pipeline.

Focusing on the motion estimation, Agrawal et al. \cite{Agrawal_2012-UnwCalib} propose an 8-point algorithm to solve for the refractive interface and camera pose using the plane of refraction (POR) constraint.
Kang et al. \cite{Kang_2012-twoViewRefractiveSfMeccv} present a two-view reconstruction approach for cameras under thin planar interfaces. 
Jordt-Sedlazeck et al. \cite{Jordt-Sedlazeck_2013RefSfM} introduce an alternating, iterative-based method for both absolute and relative pose estimation. 
Chadebecq et al. \cite{chadebecq2017refractive,chadebecq2019refractive} derive a refractive fundamental constraint for iterative refinement, mainly targeting thin flat-ports.
However, in SfM, correspondences are often contaminated by outliers, necessitating robust estimation techniques such as RANSAC \cite{Fischler1981_RandomSampling}.
The aforementioned methods are not minimal solvers, but require a good initialization and are computationally slow when using RANSAC.
Elnashef et al. \cite{elnashef2019direct} propose a linear approach to the absolute pose problem under flat refraction using the varifocal model. 
They later address relative pose estimation with a 3-point algorithm, highlighting the possibility to estimate the true scene scale \cite{elnashef2023three}.
However, determining the relative rotation requires performing a non-linear optimization to minimize the epipolar curve distances. 
While much attention has been given to flat-port cameras, little work has focused on decentered dome-ports. 
Hu et al. \cite{hu2023refractive} employ a virtual perspective camera similar to the varifocal model, proposing pose refinement methods applicable to generalized refractive camera setups.
They additionally propose a minimal solution to the relative pose estimation problem, requiring 17 point correspondences.
However, we show in our evaluation that the algorithm can only be applied underwater in very low noise conditions.
In more generalized cases, a minimal 3-point solution to absolute pose estimation is presented in \cite{hee2016minimal}, and Kneip et al. \cite{kneip2014efficient} propose an 8-point algorithm for solving the relative pose problem. 
Both approaches are designed for multi-camera systems in self-driving car scenarios, assuming relatively large baselines between the cameras and various directions they face. 
However, in the underwater-refraction induced axial camera model, the baselines between multiple virtual projection centers are small, typically in the millimeter range, which is a scenario where these approaches have not yet been tested.

Maybe surprisingly, we show that the former algorithm achieves comparable performance against the baseline, which uses standard Perspective-n-Point (PnP) algorithms \cite{gao2003complete,lepetit2009epnp} on un-refracted data, whereas the latter approach is found to be inapplicable.
Hence, we choose the 3-point algorithm for absolute pose estimation in our RSfM pipeline. 
Regarding the relative pose estimation problem, we have not found a satisfactory approach so far. 
Therefore, we propose a more practical approach that is more robust for geometric verification and SfM initialization., which will be elaborated in the next section.


\begin{figure}
	\centering
	\small
	\def\svgwidth{0.42\columnwidth}
	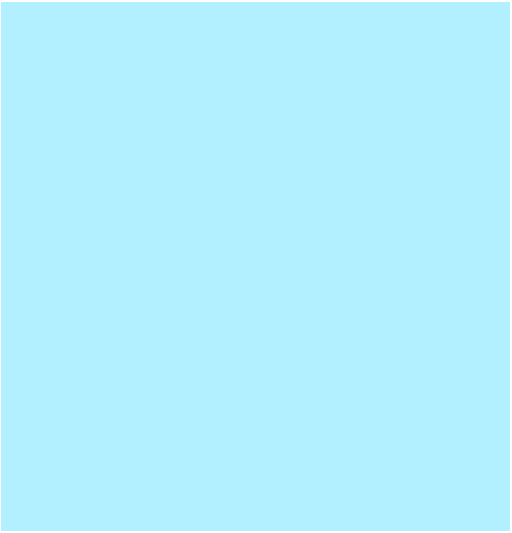
	\def\svgwidth{0.42\columnwidth}
	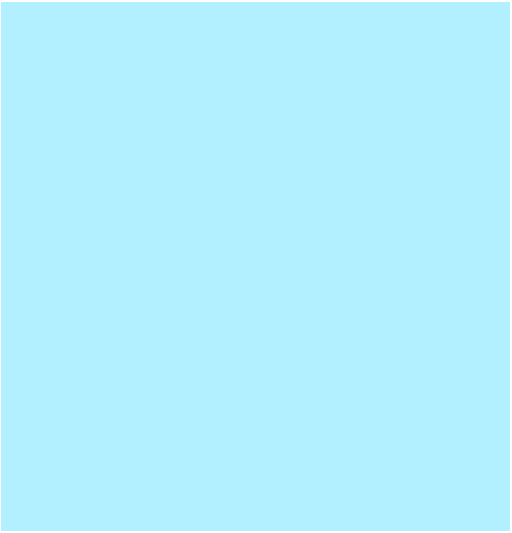
	\caption{A schematic illustration of the refractive camera models. The scene points $\d X$ are observed by the camera at image points $\d x$ through the interface. The virtual cameras $V$ are depicted by differently colored dashed triangles situated along the refraction axis $A$. \textbf{Left}: Flat-port. \textbf{Right}: Dome-port.}
	\label{fig:refrac_camera}
\end{figure}

\begin{figure}
	\centering
	\small
	\def\svgwidth{0.8\columnwidth}
	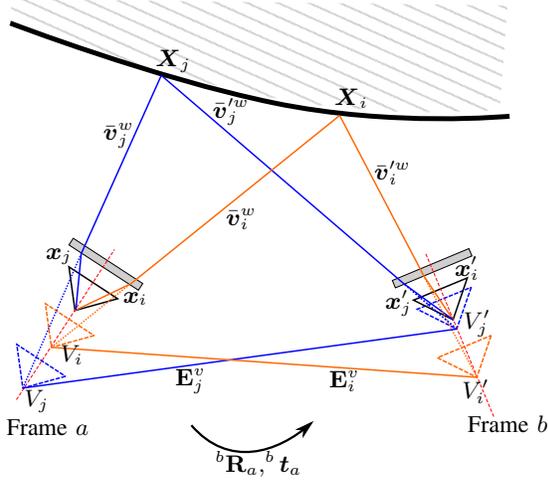
	\caption{A schematic illustration of the feature-dependent virtual epipolar geometry $\mq E^v$, which relates the relative pose $^b\mq R_{a}, ^b\d t_a$ of two frames.}
	\label{fig:virutal_epipolar}
\end{figure}

\section{Refractive Structure-from-Motion} 


\textbf{Refractive Camera Models.} To integrate refraction into the SfM process, we first make the following consideration: The refractive camera model should be generalizable to both thin/thick flat-port and dome-port, and extendable for potentially more scenarios; Additionally, the real physical camera, which is situated behind the refractive interface, should be interchangeable.
A schematic illustration of the refractive imaging setup is depicted in Fig. \ref{fig:refrac_camera}.

The real camera which is described by its intrinsic parameters $\mathcal{P}_{\mathrm{cam}}$, observes the scene points $\d X$ from an image point $\d x$ through a glass interface.
The flat-port interface is defined by parameters including the unit normal vector of the interface $\d n_{\mathrm{int}} = (n_x, n_y, n_z)\trans$, the camera-to-interface distance $d_{\mathrm{int}}$, and the thickness.
These interface parameters are defined locally relative to the camera, with $\d n_{\mathrm{int}} = (0, 0, 1)\trans$ coinciding with the optical axis of the camera.
The dome-port is characterized by its dome center (or decentering) $\d C_d = (C_x, C_y, C_z)\trans$ in the local camera coordinate frame, along with the radius and thickness.
The refraction axis $A$ describes the camera ray passing through the interface perpendicularly.
In the case of a flat-port, the refraction axis aligns with the interface normal $\d n_\mathrm{int}$, while in the case of a dome-port, it aligns with the normalized decentering direction.

According to Snell's law, the refracted normalized ray vector $\bar{\d v}_{\mathrm{refrac}}$ can be computed by:
\begin{equation}
\bar{\d v}_{\mathrm{refrac}} = r \cdot \bar{\d v} - (rc - \sqrt{1 - r^2 (1 - c^2) })\cdot \d n
\end{equation}
where $\bar{\d v}$ is the normalized incident ray, $c=\d n \cdot \bar{\d v}$ and $r = n_1 / n_2$ which represents the ratio of the two involved media's refraction indices.

In our convention, the normal vector $\d n$ points from the surface towards the side where the ray is refracted. 
We then utilize the ray-tracing technique to obtain the refracted ray in water $\d v^w$ starting from the outer interface.
The implementation of the ray-plane/sphere intersection can be found in \cite{pharr2023physically}.

\textbf{Virtual Camera Computation.} Afterwards, we replace individual rays of the refractive camera with virtual pinhole cameras (depicted as dashed triangles in Fig. \ref{fig:refrac_camera}).
These virtual cameras are feature-dependent and are positioned at the intersection of the refraction axis $A$ with $\bar{\d v}^w$, observing the same ray in water $\d v^w$ perspectively.
The pose of the virtual camera is described by a rigid transformation from the real to the virtual coordinate frame\footnote{Throughout this work, a rigid transformation $^b\mq T_a$ transforms a point in the $a$ coordinate frame to the $b$ coordinate frame.} $^v\mq T_r = (^v\mq R_r \;\vert\; ^v\d t_r)$.
This technique is initially introduced in \cite{Telem_2010-underwaterCalib} for calibration and widely employed in \cite{Jordt-Sedlazeck_2013RefSfM,elnashef2019direct,seegraberunderwater}.

However, unlike previous works where they align the virtual cameras with the refraction axis $A$ and utilize the camera-to-interface distance $d$ as the virtual focal length \cite{Jordt-Sedlazeck_2013RefSfM}, we have discovered that this approach introduces instability in the forward projection of a 3D point onto the image plane refractively when the axis $A$ is significantly off from the camera's optical axis (e.g. points are located behind the virtual camera). 
This situation can occur if the flat-port interface is severely tilted (which is unrealistic in the underwater imaging scenario), or if the decentering direction is oriented sideways in the case of a dome-port (which occurs more frequently). 
Furthermore, drastic changes in the virtual focal length can potentially lead to variations in the magnitude of the reprojection error, thereby introducing imbalance in the bundle adjustment process.
We therefore suggest to keep the rotation of the virtual camera as identity $^v\mq R_r = \mq I$, and then take the mean focal length of the real camera as the virtual focal length $f_v = f_\mathrm{mean}$. 
In addition, we determine the virtual principal points $(c_{vx}, c_{vy})$ in a way such that the original observed image point $\d x = (x, y)\trans$ remains the same:
\begin{equation}
	c_{vx} = x - f_v \cdot \bar{\q v}^w_\mathrm{hnorm}(x),\;\;c_{vy} = y - f_v \cdot \bar{\q v}^w_\mathrm{hnorm}(y)
\end{equation}
Here, $\bar{\q v}^w_\mathrm{hnorm}(x)$ and $\bar{\q v}^w_\mathrm{hnorm}(y)$ represent the $x$-, and $y$-component of the homogeneous-normalized ray in water $\bar{\d v}^w$.
Finally, the virtual camera center $^r\d t_v$ can be found by intersecting $\bar{\d v}^w$ with $A$.

\textbf{Absolute Pose Estimation.} \label{sec:motion_estim} For absolute pose estimation, we utilize the generalized absolute pose estimator (referred to as GP3P), which is readily available in COLMAP \cite{hee2016minimal}. We construct a set of virtual cameras $\mathcal{V}_\mathrm{cam}$ from a set of image points and treat them as a rigidly mounted multi-camera rig. Estimating the absolute pose of the rig is equivalent to estimating the pose of the real camera.
The algorithm requires minimally 3 point correspondences.

\textbf{Relative Pose Estimation.} The literature review has highlighted the inherent difficulty of relative pose estimation. 
In response, we propose a simplification strategy that involves a slight trade-off in accuracy. 
Rather than directly estimating the refractive relative pose, we opt to estimate the relative pose of the best-approximated perspective camera using the well-established 5-point algorithm \cite{Nister2004-Efficient}.
To compute the best-approximated perspective camera model, we randomly sample 1000 image points and back-project them to 3D space at a distance of $5m$ using the original refractive camera model.
The parameters are determined by minimizing the reprojection error of the 3D-2D points, but with the perspective camera model:
\begin{equation}
	\mathcal{P}_\mathrm{prox} = \underset{	\mathcal{P}_\mathrm{prox}}{\arg\min} \sum_{i} \Vert \pi (	\mathcal{P}_\mathrm{prox}, \d X_i) - \d x_i \Vert^2_2
\end{equation} 
where $\mathcal{P}_\mathrm{prox}$ denotes the parameters of the best-approximated camera, and $\pi (\cdot)$ represents the forward projection function.
Such an approximation will never yield a perfect solution even under noise-free, outlier-free conditions, except in the case of a perfectly centered dome-port scenario. 
However, experimental results demonstrate that it generally performs adequately, with only a marginal loss of inlier correspondences (less than $2\%$ in the worst case).

When initializing SfM from the first image pair, we additionally refine the estimated relative pose by minimizing the refractive virtual epipolar cost, similar to the approach proposed in \cite{hu2023refractive}.
As depicted in Fig. \ref{fig:virutal_epipolar}, the refracted rays $\bar{\d v}^w$ and $\bar{\d v}^{\prime w}$, along with the vector connecting the two virtual camera centers, form an epipolar plane.
Suppose the relative pose between the image pair is expressed as $^b\mq T_a = (^b\mq R_a \;\vert\; ^b\d t_a)$, and a feature point $\d x_i$ in image $a$ is matched to the feature point $\d x^\prime_i$ in image $b$.
The transformations from the real camera to the virtual one at $\d x_i$ and $\d x^\prime_i$ are $^{v_i}\mq T_r$ and $^{v^\prime_i}\mq T_r$, respectively.
Next, the transformation from the virtual camera to its corresponding virtual camera in frame $b$ is concatenated as:
\begin{equation}
	^{v^\prime_i}\mq T_{v_i} = (^{v^\prime_i}\mq R_{v_i} \;\vert\; ^{v^\prime_i}\d t_{v_i}) = ^{v^\prime_i}\mq T_r \cdot ^b\mq T_a \cdot (^{v_i}\mq T_r)^{-1}
\end{equation}
Then, for each feature point pair $\d x_i$ and $\d x^\prime_i$, we have an epipolar constraint :
\begin{equation}
\label{eq:sampson_dist}
\hat{\q x}^{\prime \trans}_i \mq E^v_i \hat{\q x}_i = 0 \;\; \mathrm{where} \;\; \mq E^v_i = [^{v^\prime_i}\d t_{v_i}]_\times \cdot ^{v^\prime_i}\mq R_{v_i}
\end{equation}
Here, $\hat{\q x_i}$ and $\hat{\q x}^\prime_i$ are the normalized coordinates.
Finally, the optimal form of $^b\mq R_a$ and $^b\d t_a$ can be obtained by minimizing the virtual epipolar cost:
\begin{equation}
^b\mq R_{a}, ^b\d t_a = \underset{^b \mq R_a, ^b\d t_a}{\arg\min} \sum_i \Vert \hat{\q x}^{\prime \trans}_i \mq E^v_i \hat{\q x}_i \Vert^2
\end{equation}
An interesting aspect of refractive relative pose estimation is the potential to estimate the baseline length. 
However, the accuracy and reliability of scale estimation are not guaranteed across all refractive camera configurations, as reported by \cite{Jordt-Sedlazeck_2013RefSfM,hu2023refractive,elnashef2023three}.
Therefore, we allow the optimizer to refine the full 6-DoFs relative pose estimation, but we do not attempt to recover the true scale. 
This decision is based on the observation that if the scale is observable (which occurs only in extreme refraction setups and very low noise conditions), it would indicate an accurate estimation of the baseline length and, consequently, a true scaled reconstruction.
On the other hand, if the scale is not observable, it implies that there is no detriment to the final reconstruction, even if the scale is completely incorrect.

\textbf{Triangulation.} We keep the triangulation algorithm unchanged from its implementation in COLMAP, only modifying it to triangulate rays generated from their respective virtual perspective cameras.
Therefore, such modification does not introduce any side effects on performance.

\textbf{Bundle Adjustment.} In classical bundle adjustment, 3D points are projected onto the image planes, and reprojection errors are minimized. 
However, in the refractive scenario, forward projection is computationally expensive. 
It involves either solving a $12^{th}$-degree polynomial or iteratively back-projecting the currently estimated projection until the error in 3D is minimized \cite{Agrawal_2012-UnwCalib,kunz2008hemispherical}.
We therefore minimize the reprojection errors on the virtual image planes, similar to \cite{Jordt-Sedlazeck_2013RefSfM}.
The refractive cost function is as follows:
\begin{equation}
	E = \sum_{j} \rho_j(\Vert \pi_{v} (\mathcal{P}_{\mathrm{cam}}, \mathcal{P}_{\mathrm{refrac}}, ^c\mq {T}_{w}, \d X_k, \d x_j) - \d x_j \Vert^2_2)
\end{equation}
where $\pi_{v}(\cdot)$ is a function that projects a 3D point $\d X_j$ to the virtual image plane. 
This function first determines the virtual camera corresponding to the feature point $\d x_j$,
and then projects the 3D point $\d X_j$ onto the virtual image plane perspectively.
A loss function $\rho_j$ is used to potentially down-weight outliers.
Both the camera intrinsic parameters $\mathcal{P}_\mathrm{cam}$ and the refractive parameters $\mathcal{P}_\mathrm{refrac}$ can be jointly refined.
However, for numerical stability reasons, we only refine the interface normal and camera-to-interface distance in the flat-port case and the decentering in the dome-port case.

\section{Evaluations}
\begin{figure}
	\centering
	\includegraphics[width=0.8\columnwidth]{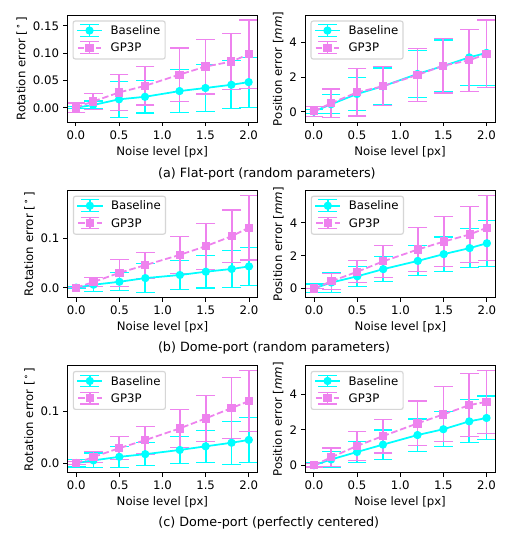}
	\caption{Numerical evaluation results of the absolute pose estimation across various refractive camera configurations.}
	\label{fig:abs_pose}
\end{figure}

\subsection{Numerical Evaluation}
Before evaluating the proposed RSfM pipeline, we first analyze the performance of the absolute and relative pose estimation, as these two steps are very critical for SfM.

\begin{figure}
	\centering
	\includegraphics[width=0.99\columnwidth]{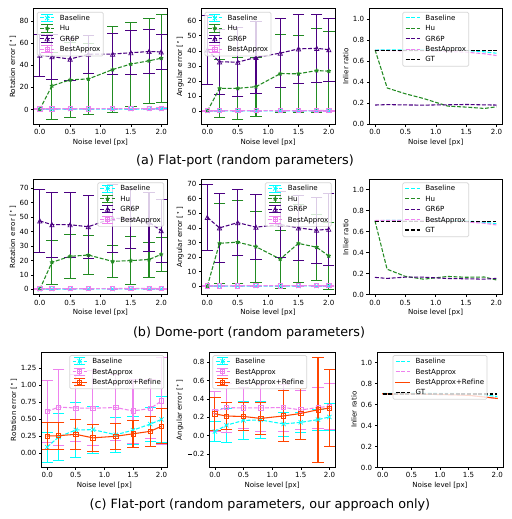}
	\caption{Numerical evaluation results of the relative pose estimation under different noise and outlier conditions.}
	\label{fig:rel_pose}
\end{figure}

\textbf{Absolute Pose Estimation.} We construct a numerical setup where a camera observes a set of randomly generated 3D points from a random pose. 
We project these points onto the image plane both with and without refraction and introduce Gaussian-distributed noise.
Additionally, a certain percentage of outliers are added to the data points. 
We then employ the GP3P method to estimate the camera pose within the RANSAC framework.
As a baseline for comparison, we perform standard PnP pose estimation \cite{gao2003complete,lepetit2009epnp} on the un-refracted data points.
The simulated camera has an image size of $1920\times1280$ pixels and a field of view of $73^\circ$. 
Each experiment consists of 200 points in total, with $30\%$ of them being outliers. 
We conduct 1000 experiments, with the refractive parameters, 2D-3D points, and camera poses randomly generated within realistic bounds, for each experiment, and measure the rotation error in degrees and position error in $mm$. 
Fig. \ref{fig:abs_pose} (a) and (b) show the flat- and dome-port setups. 
We also questioned ourselves whether the approach would become numerically unstable or degenerate in a scenario with minimal refractive effects, i.e. when the camera system is near central. To address this we conduct a third evaluation using a perfectly centered dome-port camera, shown in Fig. \ref{fig:abs_pose} (c). 
The GP3P pose estimator consistently demonstrates performance comparable to the baseline approach across various evaluation configurations. 
In nearly all cases, the GP3P exhibits only marginal differences from the baseline, with maximum deviations of less than $0.2^\circ$ in rotation error and $4mm$ in position error. 
Furthermore, the correct outlier ratio is reported by RANSAC in all cases.
Note that there is no non-linear refinement involved in this evaluation.
While the GP3P estimator was originally developed for multi-camera systems in self-driving car scenarios, we investigate its stability and performance when applied to our refractive camera setup. 
The results depicted in Fig. \ref{fig:abs_pose} (c) demonstrate that the approach remains robust and capable of handling such scenarios effectively. 
Based on these findings, we conclude that the GP3P estimator is sufficient for our application.
\begin{table*}
	\centering
	\footnotesize
	\caption{Evaluation results on re-rendered datasets. The optimal results are highlighted in \textbf{Bold} text.}
	\label{tab:eval_render}       
	\scalebox{0.7}{
		\begin{tabular}{cccccccccccccccccccccc}
			\hline\noalign{\smallskip}
			\multicolumn{2}{c}{\multirow{2}{*}{Datasets}} & \multirow{2}{*}{$N$} & \multicolumn{4}{c}{\textbf{UWPinhole}} & \multicolumn{4}{c}{\textbf{RSfM (Use GT Calib)}} & \multicolumn{4}{c}{\textbf{RSfM (Refine)}}\\
			& & & RE & $\Delta \mq R$ & $\Delta \d t$ & $\Delta d$ & RE & $\Delta \mq R$ & $\Delta \d t$ & $\Delta d$ & RE & $\Delta \mq R$ & $\Delta \d t$ & $\Delta d$\\
			\noalign{\smallskip}\hline\noalign{\smallskip}
			\multirow{8}{*}{Tank} & \textit{Flat+Ortho+Close} & 106 & 0.522 & 0.143 & 1.978 & 2.774 & \textbf{0.347} & \textbf{0.016} & \textbf{0.230} & \textbf{2.098} & 0.348 & 0.031 & 1.169 & 2.254\\
					   			  & \textit{Flat+Tilt+Close} & 106 & 1.147 & 6.442 & 19.432 & 19.060 & \textbf{0.330} & \textbf{0.013} & \textbf{0.252} & \textbf{1.657} & 0.332 & 0.027 & 1.077 & 1.844\\
								  & \textit{Flat+Ortho+Far} & 106 & 0.773 & 0.957 & 17.324 & 7.005 & \textbf{0.335} & \textbf{0.021} & \textbf{0.236} & 1.964 & \textbf{0.335} & 0.022 & 0.307 & \textbf{1.939}\\
								  & \textit{Flat+Tilt+Far} & 106 & 1.180 & 5.774 & 23.276 & 16.052 & \textbf{0.340} & \textbf{0.012} & \textbf{0.218} & \textbf{1.665} & 0.344 & 0.031 & 1.212 & 1.918\\
								  & \textit{Dome+Backward+Close} & 106 & 0.313 & 0.017 & 0.516 & 1.915 & \textbf{0.312} & \textbf{0.011} & 0.577 & 1.852 & \textbf{0.312} & 0.027 & \textbf{0.462} & \textbf{1.797}\\
								  & \textit{Dome+Backward+Far} & 106 & 0.390 & 0.181 & 6.193 & 3.854 & \textbf{0.306} & \textbf{0.006} & \textbf{0.083} & 2.225 & \textbf{0.306} & 0.013 & 0.440 & \textbf{2.195} \\
								  & \textit{Dome+Sideward+Close} & 106 & 0.312 & 0.735 & 0.510 & 1.923 & \textbf{0.306} & 0.040 & 0.551 & 1.880 & \textbf{0.306} & \textbf{0.026} & \textbf{0.227} & \textbf{1.863} \\
								  & \textit{Dome+Sideward+Far} & 106 & 1.124 & 5.247 & 6.246 & 7.595 & \textbf{0.309} & \textbf{0.009} & 0.092 & \textbf{2.081} & \textbf{0.309} & 0.012 & \textbf{0.088} & 2.088 \\
			\hline\noalign{\smallskip}
			\multirow{2}{*}{AUV}  & \textit{Flat+Ortho} & 5740 & 0.277 & 10.272 & 893.730 & 816.131 & \textbf{0.199} & \textbf{0.200} & \textbf{27.933} & \textbf{11.842} & \textbf{0.199} & 0.217 & 29.441 & 13.778 \\
								  & \textit{Flat+Tilt} & 5740 & 0.825 & - & - & - & \textbf{0.196} & \textbf{0.238} & \textbf{29.182} & 16.247 & 0.197 & 0.256 & 29.337 & \textbf{16.020}\\  
			\hline\noalign{\smallskip}
		\end{tabular}
	}
\end{table*}

\begin{table}
	\centering
	\footnotesize
	\caption{Results of the estimated refractive parameters when performing refinement in RSfM.}
	\label{tab:eval_calib_render}       
	\scalebox{0.58}{
		\begin{tabular}{cccccccccccccccccccccc}
			\hline\noalign{\smallskip}
			\multicolumn{2}{c}{Datasets} & \multicolumn{4}{c}{\textbf{GT ($n_x, n_y, n_z, d_\mathrm{int}$) }} & \multicolumn{4}{c}{\textbf{Est. ($n_x, n_y, n_z, d_\mathrm{int}$)}}\\
			\noalign{\smallskip}\hline\noalign{\smallskip}
			\multirow{4}{*}{Tank} & \textit{Flat+Ortho+Close} & 0 & 0& 1& 0.01& 8.80e-05& -3.51e-05 & 1.000& 0.035\\
			& \textit{Flat+Tilt+Close} & 0.166& 0.148& 0.975& 0.01& 0.166& 0.148& 0.975& 0.033\\
			& \textit{Flat+Ortho+Far} & 0& 0& 1& 0.05& 9.81e-05& -3.30e-05& 1.000& 0.053\\
			& \textit{Flat+Tilt+Far} & 0.166& 0.148& 0.975& 0.05& 0.166& 0.148& 0.975& 0.190\\
			\multirow{2}{*}{AUV}  & \textit{Flat+Ortho} & 0 & 0& 1& 0.02& -3.71e-05& 8.11e-05& 1.000& 0.0201\\
			& \textit{Flat+Tilt} & 0.166 & 0.148& 0.975& 0.02& 0.166& 0.148& 0.975& 0.0201\\  
            \hline\noalign{\smallskip}
            \multicolumn{2}{c}{Datasets} & \multicolumn{4}{c}{\textbf{GT ($C_x, C_y, C_z$) }} & \multicolumn{4}{c}{\textbf{Est. ($C_x, C_y, C_z$)}}\\
			\hline\noalign{\smallskip}
             \multirow{4}{*}{Tank} & \textit{Dome+Backward+Close} & 0& 0& 0.003& & 5.14e-06 & -8.14e-05& 0.003\\
			& \textit{Dome+Backward+Far} & 0& 0& 0.03& & 6.66e-06& -3.13e-05& 0.030\\
			& \textit{Dome+Sideward+Close} & 0.003& 0& 0 & & 0.003& -6.05e-05& -8.47e-05\\
			& \textit{Dome+Sideward+Far} & 0.03& 0& 0& & 0.030& -4.07e-05& -4.96e-05\\
			\hline\noalign{\smallskip}
		\end{tabular}
        }
\end{table}

\textbf{Relative Pose Estimation.} We conduct the same experiments as the previous ones to evaluate relative pose estimation, except that random 2D-2D correspondences are generated. 
Since an accurate estimation of the baseline length cannot be guaranteed, we measure rotation error in degrees and the angular error between the estimated and ground truth relative translation direction, also in degrees, and the inlier ratio reported by RANSAC.
We evaluate several minimal solvers: Hu's 17-point algorithm \cite{hu2023refractive}, Kneip's 8-point generalized relative pose solver (GR6P), and our proposed method (denoted as BestApprox). The baseline method for comparison is the 5-point algorithm \cite{Nister2004-Efficient} using un-refracted data points.
Fig. \ref{fig:rel_pose} (a) and (b) present the evaluation results for the flat-port and dome-port setups, respectively, without involving any non-linear refinement.
Fig. \ref{fig:rel_pose} (c) exclusively displays the evaluation outcomes of our approach alongside the non-linear refined results compared to the baseline in the flat-port setups.

As depicted in Fig. \ref{fig:rel_pose} (a) and (b), Hu's approach nearly recovers ground truth results under low noise conditions, but does not deliver satisfactory results when the noise level is increased, and the inlier ratio drops rapidly, similar to their report \cite{hu2023refractive}.
The GR6P algorithm performs poorly already under low noise conditions, meaning that the approach is inapplicable to the underwater-refraction induced axial camera model.
Nevertheless, our approach performs stably and robustly under various conditions, with accuracy only marginally worse than the baseline method, as evident from Fig. \ref{fig:rel_pose} (c) where the other approaches are excluded. 
The maximum loss of inliers is only less than $2\%$ in the worst case as compared to the baseline.
Furthermore, refining the initial estimated relative pose by minimizing the virtual epipolar cost further improves the accuracy, ensuring an accurate and robust RSfM initialization.

\subsection{Re-Render from Real-World}

\begin{figure}
    \centering
    \subfloat[Tank Scene]{
        \includegraphics[width=0.45\columnwidth]{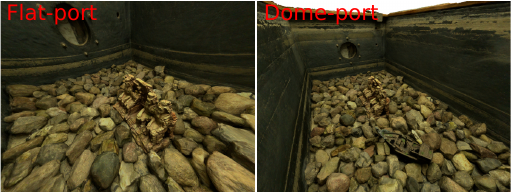}
    }
    \subfloat[AUV Scene]{
        \includegraphics[width=0.45\columnwidth]{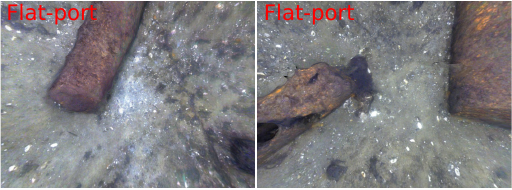}
    }
    \caption{Example images of the re-rendered refractive datasets.}
    \label{fig:example_rerender}
\end{figure}

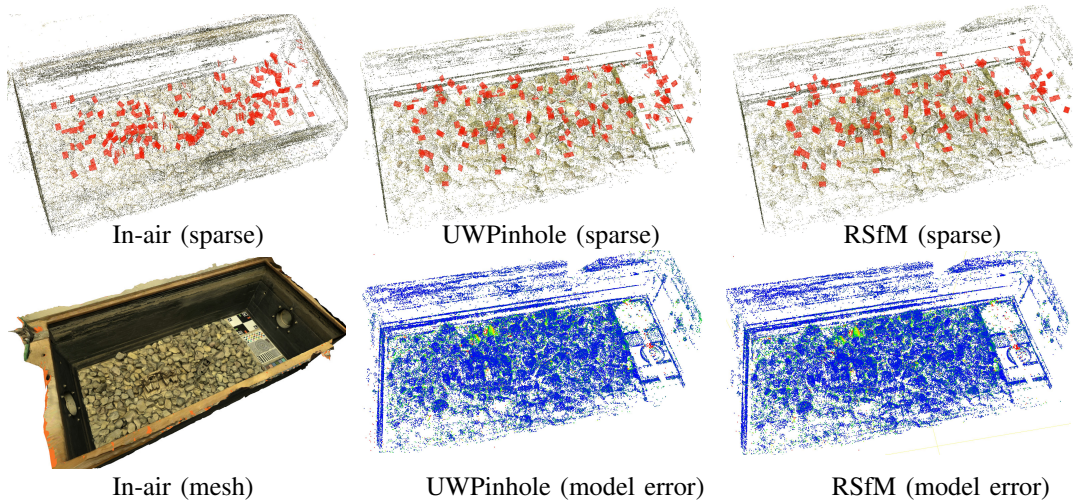
\begin{figure*}
    \centering
    \def\svgwidth{0.8\textwidth}
\begingroup%
  \makeatletter%
  \providecommand\color[2][]{%
    \errmessage{(Inkscape) Color is used for the text in Inkscape, but the package 'color.sty' is not loaded}%
    \renewcommand\color[2][]{}%
  }%
  \providecommand\transparent[1]{%
    \errmessage{(Inkscape) Transparency is used (non-zero) for the text in Inkscape, but the package 'transparent.sty' is not loaded}%
    \renewcommand\transparent[1]{}%
  }%
  \providecommand\rotatebox[2]{#2}%
  \newcommand*\fsize{\dimexpr\f@size pt\relax}%
  \newcommand*\lineheight[1]{\fontsize{\fsize}{#1\fsize}\selectfont}%
  \ifx\svgwidth\undefined%
    \setlength{\unitlength}{505.89001465bp}%
    \ifx\svgscale\undefined%
      \relax%
    \else%
      \setlength{\unitlength}{\unitlength * \real{\svgscale}}%
    \fi%
  \else%
    \setlength{\unitlength}{\svgwidth}%
  \fi%
  \global\let\svgwidth\undefined%
  \global\let\svgscale\undefined%
  \makeatother%
  \begin{picture}(1,0.47441142)%
    \lineheight{1}%
    \setlength\tabcolsep{0pt}%
    \put(0,0){\includegraphics[width=\unitlength,page=1]{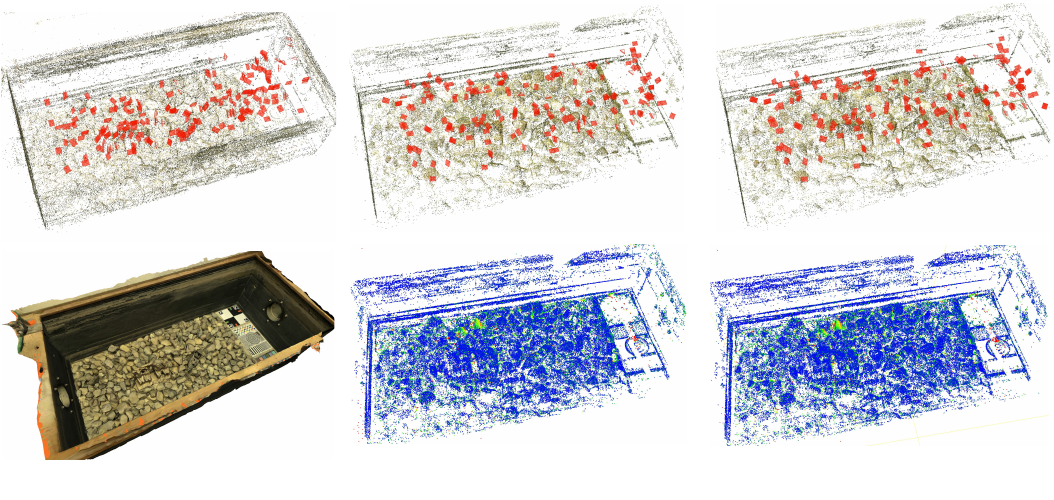}}%
    \put(0.09940747,0.2494719){\color[rgb]{0,0,0}\makebox(0,0)[lt]{\lineheight{1.25}\smash{\begin{tabular}[t]{l}In-air (sparse)\end{tabular}}}}%
    \put(0.40758812,0.2494719){\color[rgb]{0,0,0}\makebox(0,0)[lt]{\lineheight{1.25}\smash{\begin{tabular}[t]{l}UWPinhole (sparse)\end{tabular}}}}%
    \put(0.78300822,0.2494719){\color[rgb]{0,0,0}\makebox(0,0)[lt]{\lineheight{1.25}\smash{\begin{tabular}[t]{l}RSfM (sparse)\end{tabular}}}}%
    \put(0.77180165,0.01413395){\color[rgb]{0,0,0}\makebox(0,0)[lt]{\lineheight{1.25}\smash{\begin{tabular}[t]{l}RSfM (model error)\end{tabular}}}}%
    \put(0.39263779,0.01413395){\color[rgb]{0,0,0}\makebox(0,0)[lt]{\lineheight{1.25}\smash{\begin{tabular}[t]{l}UWPinhole (model error)\end{tabular}}}}%
    \put(0.09940747,0.01413395){\color[rgb]{0,0,0}\makebox(0,0)[lt]{\lineheight{1.25}\smash{\begin{tabular}[t]{l}In-air (mesh)\end{tabular}}}}%
  \end{picture}%
\endgroup%

    \caption{Reconstruction results of the real water tank dataset.}
    \label{fig:eval_real_tank}
\end{figure*}
To benchmark our proposed RSfM approach, we render novel refractive images from 3D meshes reconstructed out of an existing refraction-free real-world dataset using a physically-based ray-tracer. We maintain identical camera poses during re-rendering to ensure a faithful emulation of the original photographic missions. The refractive effects are simulated during rendering by digitally placing a glass-material interface in front of the camera.

To evaluate the robustness of the system, we render the same scene with various refractive camera setups. 
These setups include orthogonal and tilted flat-port interfaces, as well as variations in the camera-to-interface distance and dome-port decentering.
The parameters we use for re-rendering are shown in Tab. \ref{tab:eval_calib_render}.
The first scene for re-rendering is a well-decorated test tank without water, reconstructed by COLMAP using a GoPro9 camera in-air under homogeneous illumination.
The second scene contains a large-scale 3D reconstruction of seafloor (scene size $44m \times 35m$) obtained using the method described in \cite{she2023semihierarchical}.
The original images of this dataset were acquired by an AUV equipped with a calibrated dome-port camera system in a real-world mapping mission.
This is to demonstrate the applicability of our approach on real-world AUV-based large-scale refractive seafloor reconstruction.
Example images of the scenes are shown in Fig. \ref{fig:example_rerender}.


We evaluate our system in three runs, always initializing the intrinsics with the ground truth in-air calibration. Specifically, we compare 1) UWPinhole which refines the intrinsics and distortion parameters; 2) RSfM using ground truth refractive parameters and keeping them constant; 3) RSfM using incorrect refractive parameters as initialization, and only refining them during bundle adjustment.
For the AUV scene, we set the baseline of the initial registered two images as the baseline measured by the navigation data to constrain the true scene scale.
Nevertheless, the navigation data is not used for RSfM reconstruction.
The results are presented in Tab. \ref{tab:eval_render}, where RE stands for the reprojection error in pixels. 
$\Delta \mq R$, $\Delta \mq t$ represent the rotation error in degrees, position errors in $mm$ respectively. All error measures are averaged across all images.
The 3D model error $\Delta d$ is measured as the average closest distance of the reconstructed sparse point cloud to the 3D mesh from which images are rendered in $mm$.
In addition, Tab. \ref{tab:eval_calib_render} shows the estimated refractive parameters against the ground truth values when refining the parameters in bundle adjustment, where the optimizable parameters are highlighted in \textbf{Bold} text.

It is evident from Tab. \ref{tab:eval_render} that our proposed RSfM approach consistently yields the best results in terms of the accuracy of both camera poses and the 3D model across various refractive camera configurations.
However, it is also interesting to note that absorbing refraction in distortion parameters is not necessarily a bad practice in some scenarios. 
For instance, when the flat-port interface is orthogonal, the refraction effects are mostly symmetric, and radial distortion can effectively absorb these effects for reasonable distance ranges. 
Similarly, in the case of a slightly decentered dome-port system where the refraction effects are not pronounced, ignoring refraction can still yield decent results. 
However, when dealing with large datasets, such as in the AUV mapping scenario, the limitations of the UWPinhole approach become apparent. 
As shown in Fig. \ref{fig:rerender_auv} (top), despite achieving a low reprojection error of only 0.277 pixels, the UWPinhole approach leads to a severely distorted reconstruction.
This occurs even when the interface normal is orthogonal, and the approach fails to produce meaningful results when the normal is tilted.
In contrast, the proposed RSfM approach can handle various dataset types and camera configurations effectively. 

\textbf{Scale Awareness.} Tab. \ref{tab:eval_calib_render} demonstrates that refining refractive parameters in RSfM effectively recovers incorrectly initialized values except for the camera-to-interface distance $d_\mathrm{int}$, which is only estimated up to scale.
This observation also aligns with the findings of \cite{elnashef2022target}, because scaling the entire scene, including $d_\mathrm{int}$, does not alter the angle of incident rays at the interface. 
Therefore, constraining the scene scale can lead to true $d_\mathrm{int}$ calibration as evident from the AUV scene results where external information such as navigation data is utilized to initialize the scene scale, resulting in the calibration of $d_\mathrm{int}$ to its true value.
\subsection{Real-World Experiments}

To obtain ground truth for evaluating the RSfM approach, we employ a GoPro9 camera to perform an in-air scan of the decorated tank without water using standard COLMAP. 
An AruCo checkerboard is positioned on the floor as a reference target for alignment based on similarity transforms, and the resulting in-air scanned model is considered the ground truth. 
The in-air reconstruction of the tank is shown in Fig. \ref{fig:eval_real_tank} (left).
Subsequently, the tank is filled with water, and an underwater dataset is captured by the same GoPro camera with a flat-port case. 
The flat-port parameters are obtained through underwater calibration \cite{JordtSedlazeck_2012RefCalibUnw}.
The acquired images have dimensions of $5184\times3888$ pixels. 
Similar to previous experiments, we reconstruct the model once using the UWPinhole approach and once with our RSfM approach.
A view of the reconstructed results are presented in Fig. \ref{fig:eval_real_tank} (center and right).
The reprojection errors of the reconstructions are 1.109 pixels for the in-air scan, 1.050 pixels using the UWPinhole approach, and 1.037 pixels with the RSfM approach.
In addition, all reconstructions are aligned, and the model error is measured as the cloud-to-mesh distance using CloudCompare. 
The model error for the UWPinhole reconstruction is $2.061mm$, and for the RSfM approach, it is $2.103mm$. 
This experiment demonstrates that the RSfM approach can achieve ground truth-level reconstruction even without accurate measures of the refractive interface. However, in this specific setup where the flat-port case for the GoPro camera is only around $2mm$ thick and the camera-to-interface distance is even less than $2mm$, there is no clear advantage over simply ignoring refraction and reconstructing using standard COLMAP. 
Additionally, the relatively small scene size of about $2m \times 1m$ and small altitude variations may not fully exploit the advantages of RSfM.
Nonetheless, Fig. \ref{fig:rerender_auv} demonstrates the necessity of the RSfM approach when mapping a large area of the seafloor using a refractive camera. 
Therefore, we present this approach to the community for situations where considering refraction is necessary.

\section{CONCLUSIONS}
We have introduced a comprehensive refractive Structure-from-Motion (RSfM) pipeline for underwater 3D reconstruction, which has been integrated into the widely used open-source SfM framework COLMAP. Our proposed components enable robust and accurate geometric verification and SfM initialization. Through comprehensive evaluations, we have demonstrated the accuracy and robustness of each individual component as well as the overall system performance. Our implementation is publicly available as an underwater extension of COLMAP.

\addtolength{\textheight}{-5cm}   








\bibliographystyle{IEEEtran}
\bibliography{reference}

\begin{thebibliography}{10}
\providecommand{\url}[1]{#1}
\csname url@rmstyle\endcsname
\providecommand{\newblock}{\relax}
\providecommand{\bibinfo}[2]{#2}
\providecommand\BIBentrySTDinterwordspacing{\spaceskip=0pt\relax}
\providecommand\BIBentryALTinterwordstretchfactor{4}
\providecommand\BIBentryALTinterwordspacing{\spaceskip=\fontdimen2\font plus
\BIBentryALTinterwordstretchfactor\fontdimen3\font minus \fontdimen4\font\relax}
\providecommand\BIBforeignlanguage[2]{{%
\expandafter\ifx\csname l@#1\endcsname\relax
\typeout{** WARNING: IEEEtran.bst: No hyphenation pattern has been}%
\typeout{** loaded for the language `#1'. Using the pattern for}%
\typeout{** the default language instead.}%
\else
\language=\csname l@#1\endcsname
\fi
#2}}

\bibitem{shortis2015_unwcalib}
\BIBentryALTinterwordspacing
M.~Shortis, ``Calibration techniques for accurate measurements by underwater camera systems,'' \emph{Sensors}, vol.~15, no.~12, pp. 30\,810--30\,826, 2015. [Online]. Available: \url{http://www.mdpi.com/1424-8220/15/12/29831}
\BIBentrySTDinterwordspacing

\bibitem{schonberger2016structure}
J.~L. Schonberger and J.-M. Frahm, ``Structure-from-motion revisited,'' in \emph{Proceedings of the IEEE Conference on Computer Vision and Pattern Recognition}, 2016, pp. 4104--4113.

\bibitem{rofallski2022investigating}
R.~Rofallski, O.~Kahmen, and T.~Luhmann, ``Investigating distance-dependent distortion in multimedia photogrammetry for flat refractive interfaces,'' \emph{The International Archives of the Photogrammetry, Remote Sensing and Spatial Information Sciences}, vol.~48, pp. 127--134, 2022.

\bibitem{treibitz_08-flatRefractive}
T.~Treibitz, Y.~Y. Schechner, and H.~Singh, ``Flat refractive geometry,'' in \emph{Proc. IEEE Conference on Computer Vision and Pattern Recognition CVPR 2008}, 2008, pp. 1--8.

\bibitem{Telem_2010-underwaterCalib}
\BIBentryALTinterwordspacing
G.~Telem and S.~Filin, ``Photogrammetric modeling of underwater environments,'' \emph{ISPRS Journal of Photogrammetry and Remote Sensing}, vol.~65, no.~5, pp. 433--444, 2010. [Online]. Available: \url{http://www.sciencedirect.com/science/article/B6VF4-50F9H66-1/2/d8dba566f79b0a207e13a6aa2bf3f69d}
\BIBentrySTDinterwordspacing

\bibitem{jordt_2016_refractive}
\BIBentryALTinterwordspacing
A.~Jordt, K.~K{\"o}ser, and R.~Koch, ``Refractive 3d reconstruction on underwater images,'' \emph{Methods in Oceanography}, vol. 15-16, pp. 90 -- 113, 2016. [Online]. Available: \url{http://www.sciencedirect.com/science/article/pii/S2211122015300086}
\BIBentrySTDinterwordspacing

\bibitem{she2021refractive}
\BIBentryALTinterwordspacing
M.~She, D.~Nakath, Y.~Song, and K.~K{\"o}ser, ``Refractive geometry for underwater domes,'' \emph{ISPRS Journal of Photogrammetry and Remote Sensing}, vol. 183, pp. 525--540, 2022. [Online]. Available: \url{https://www.sciencedirect.com/science/article/pii/S092427162100304X}
\BIBentrySTDinterwordspacing

\bibitem{agrawal2010analytical}
A.~Agrawal, Y.~Taguchi, and S.~Ramalingam, ``Analytical forward projection for axial non-central dioptric and catadioptric cameras,'' in \emph{European Conference on Computer Vision}.\hskip 1em plus 0.5em minus 0.4em\relax Springer, 2010, pp. 129--143.

\bibitem{JordtSedlazeck_2012RefCalibUnw}
A.~Jordt-Sedlazeck and R.~Koch, ``Refractive calibration of underwater cameras,'' in \emph{Computer Vision - ECCV 2012}, ser. Lecture Notes in Computer Science, A.~Fitzgibbon, S.~Lazebnik, P.~Pietro, Y.~Sato, and C.~Schmid, Eds.\hskip 1em plus 0.5em minus 0.4em\relax Springer Berlin Heidelberg, 2012, vol. 7576, pp. 846--859.

\bibitem{she2019adjustment}
M.~She, Y.~Song, J.~Mohrmann, and K.~K{\"o}ser, ``Adjustment and calibration of dome port camera systems for underwater vision,'' in \emph{German Conference on Pattern Recognition}.\hskip 1em plus 0.5em minus 0.4em\relax Springer, 2019, pp. 79--92.

\bibitem{chadebecq2019refractive}
F.~Chadebecq, F.~Vasconcelos, R.~Lacher, E.~Maneas, A.~Desjardins, S.~Ourselin, T.~Vercauteren, and D.~Stoyanov, ``Refractive two-view reconstruction for underwater 3d vision,'' \emph{International Journal of Computer Vision}, pp. 1--17, 2019.

\bibitem{elnashef2023three}
B.~Elnashef and S.~Filin, ``A three-point solution with scale estimation ability for two-view flat-refractive underwater photogrammetry,'' \emph{ISPRS Journal of Photogrammetry and Remote Sensing}, vol. 198, pp. 223--237, 2023.

\bibitem{Li_08GeneralEEstHartley}
H.~Li, R.~Hartley, and J.-H. Kim, ``A linear approach to motion estimation using generalized camera models,'' in \emph{Computer Vision and Pattern Recognition, 2008. CVPR 2008. IEEE Conference on}, 7 2008, pp. 1 --8.

\bibitem{hu2023refractive}
X.~Hu, F.~Lauze, and K.~S. Pedersen, ``Refractive pose refinement: Generalising the geometric relation between camera and refractive interface,'' \emph{International Journal of Computer Vision}, vol. 131, no.~6, pp. 1448--1476, 2023.

\bibitem{Jordt-Sedlazeck_2013RefSfM}
A.~Jordt-Sedlazeck and R.~Koch, ``Refractive structure-from-motion on underwater images,'' in \emph{Computer Vision (ICCV), 2011 IEEE International Conference on}, 2013, pp. 57--64.

\bibitem{chadebecq2017refractive}
F.~Chadebecq, F.~Vasconcelos, G.~Dwyer, R.~Lacher, S.~Ourselin, T.~Vercauteren, and D.~Stoyanov, ``Refractive structure-from-motion through a flat refractive interface,'' in \emph{Proceedings of the IEEE International Conference on Computer Vision}, 2017, pp. 5315--5323.

\bibitem{elnashef2023drift}
B.~Elnashef and S.~Filin, ``Drift reduction in underwater egomotion computation by axial camera modeling,'' \emph{IEEE Robotics and Automation Letters}, 2023.

\bibitem{schonberger2016pixelwise}
J.~L. Sch{\"o}nberger, E.~Zheng, J.-M. Frahm, and M.~Pollefeys, ``Pixelwise view selection for unstructured multi-view stereo,'' in \emph{European Conference on Computer Vision}.\hskip 1em plus 0.5em minus 0.4em\relax Springer, 2016, pp. 501--518.

\bibitem{mildenhall2021nerf}
B.~Mildenhall, P.~P. Srinivasan, M.~Tancik, J.~T. Barron, R.~Ramamoorthi, and R.~Ng, ``Nerf: Representing scenes as neural radiance fields for view synthesis,'' \emph{Communications of the ACM}, vol.~65, no.~1, pp. 99--106, 2021.

\bibitem{billings2022hybrid}
G.~Billings, R.~Camilli, and M.~Johnson-Roberson, ``Hybrid visual slam for underwater vehicle manipulator systems,'' \emph{IEEE Robotics and Automation Letters}, vol.~7, no.~3, pp. 6798--6805, 2022.

\bibitem{Grossberg_05-RaxelImagingModel}
M.~D. Grossberg and S.~K. Nayar, ``The raxel imaging model and ray-based calibration,'' \emph{International Journal of Computer Vision}, vol.~61, no.~2, pp. 119--137, 2005.

\bibitem{schops2020having}
T.~Schops, V.~Larsson, M.~Pollefeys, and T.~Sattler, ``Why having 10,000 parameters in your camera model is better than twelve,'' in \emph{Proceedings of the IEEE/CVF Conference on Computer Vision and Pattern Recognition}, 2020, pp. 2535--2544.

\bibitem{Agrawal_2012-UnwCalib}
A.~Agrawal, S.~Ramalingam, Y.~Taguchi, and V.~Chari, ``A theory of multi-layer flat refractive geometry,'' in \emph{CVPR}, 2012.

\bibitem{kunz2008hemispherical}
C.~Kunz and H.~Singh, ``Hemispherical refraction and camera calibration in underwater vision,'' in \emph{OCEANS 2008}.\hskip 1em plus 0.5em minus 0.4em\relax IEEE, 2008, pp. 1--7.

\bibitem{LUCZYNSKI20179}
\BIBentryALTinterwordspacing
T.~Luczynski, M.~Pfingsthorn, and A.~Birk, ``The pinax-model for accurate and efficient refraction correction of underwater cameras in flat-pane housings,'' \emph{Ocean Engineering}, vol. 133, pp. 9 -- 22, 2017. [Online]. Available: \url{http://www.sciencedirect.com/science/article/pii/S0029801817300434}
\BIBentrySTDinterwordspacing

\bibitem{sturm2005multi}
P.~Sturm, ``Multi-view geometry for general camera models,'' in \emph{2005 IEEE Computer Society Conference on Computer Vision and Pattern Recognition (CVPR'05)}, vol.~1.\hskip 1em plus 0.5em minus 0.4em\relax IEEE, 2005, pp. 206--212.

\bibitem{sturm_06-GenericCameras}
P.~Sturm, S.~Ramalingam, and S.~Lodha, ``On calibration, structure from motion and multi-view geometry for generic camera models,'' in \emph{Imaging Beyond the Pinhole Camera}, ser. Computational Imaging and Vision, K.~Daniilidis and R.~Klette, Eds.\hskip 1em plus 0.5em minus 0.4em\relax Springer, aug 2006, vol.~33.

\bibitem{Ramalingam_2006genericSfMFramework}
S.~Ramalingam, S.~K. Lodha, and P.~Sturm, ``A generic structure-from-motion framework,'' \emph{Computer Vision and Image Understanding}, vol. 103, no.~3, pp. 218--228, Sept. 2006.

\bibitem{chari2009refractiveplane}
\BIBentryALTinterwordspacing
V.~Chari and P.~Sturm, ``{Multiple-View Geometry of the Refractive Plane},'' in \emph{{BMVC 2009 - 20th British Machine Vision Conference}}, A.~Cavallaro, S.~Prince, and D.~C. Alexander, Eds.\hskip 1em plus 0.5em minus 0.4em\relax London, United Kingdom: {The British Machine Vision Association (BMVA)}, Sept. 2009, pp. 1--11. [Online]. Available: \url{https://hal.inria.fr/inria-00434342}
\BIBentrySTDinterwordspacing

\bibitem{Kang_2012-twoViewRefractiveSfMeccv}
L.~Kang, L.~Wu, and Y.-H. Yang, ``Two-view underwater structure and motion for cameras under flat refractive interfaces,'' in \emph{Computer Vision - ECCV 2012}, ser. Lecture Notes in Computer Science, A.~Fitzgibbon, S.~Lazebnik, P.~Perona, Y.~Sato, and C.~Schmid, Eds.\hskip 1em plus 0.5em minus 0.4em\relax Springer Berlin / Heidelberg, 2012, vol. 7575, pp. 303--316.

\bibitem{Fischler1981_RandomSampling}
M.~Fischler and R.~Bolles, ``{RAN}dom {SA}mpling {C}onsensus: a paradigm for model fitting with application to image analysis and automated cartography,'' \emph{Communications of the ACM}, vol.~24, no.~6, pp. 381--395, 6 1981.

\bibitem{elnashef2019direct}
B.~Elnashef and S.~Filin, ``Direct linear and refraction-invariant pose estimation and calibration model for underwater imaging,'' \emph{ISPRS Journal of Photogrammetry and Remote Sensing}, vol. 154, pp. 259--271, 2019.

\bibitem{hee2016minimal}
G.~Hee~Lee, B.~Li, M.~Pollefeys, and F.~Fraundorfer, ``Minimal solutions for pose estimation of a multi-camera system,'' in \emph{Robotics Research: The 16th International Symposium ISRR}.\hskip 1em plus 0.5em minus 0.4em\relax Springer, 2016, pp. 521--538.

\bibitem{kneip2014efficient}
L.~Kneip and H.~Li, ``Efficient computation of relative pose for multi-camera systems,'' in \emph{Proceedings of the IEEE conference on computer vision and pattern recognition}, 2014, pp. 446--453.

\bibitem{gao2003complete}
X.-S. Gao, X.-R. Hou, J.~Tang, and H.-F. Cheng, ``Complete solution classification for the perspective-three-point problem,'' \emph{IEEE transactions on pattern analysis and machine intelligence}, vol.~25, no.~8, pp. 930--943, 2003.

\bibitem{lepetit2009epnp}
V.~Lepetit, F.~Moreno-Noguer, and P.~Fua, ``Epnp: An accurate o (n) solution to the pnp problem,'' \emph{International journal of computer vision}, vol.~81, no.~2, p. 155, 2009.

\bibitem{pharr2023physically}
M.~Pharr, W.~Jakob, and G.~Humphreys, \emph{Physically based rendering: From theory to implementation}.\hskip 1em plus 0.5em minus 0.4em\relax MIT Press, 2023.

\bibitem{seegraberunderwater}
F.~Seegr{\"a}ber, P.~Sch{\"o}ntag, F.~Woelk, and K.~K{\"o}ser, ``Underwater multiview stereo using axial camera models.''

\bibitem{Nister2004-Efficient}
D.~Nist{\'e}r, ``An efficient solution to the five-point relative pose problem,'' \emph{TPAMI}, vol.~26, pp. 756--777, 2004.

\bibitem{she2023semihierarchical}
M.~She, Y.~Song, D.~Nakath, and K.~K{\"o}ser, ``Semihierarchical reconstruction and weak-area revisiting for robotic visual seafloor mapping,'' \emph{Journal of Field Robotics}, 2023.

\bibitem{elnashef2022target}
B.~Elnashef and S.~Filin, ``Target-free calibration of flat refractive imaging systems using two-view geometry,'' \emph{Optics and Lasers in Engineering}, vol. 150, p. 106856, 2022.

\end{thebibliography}

\end{document}